\documentclass{sig-alternate-05-2015}

\usepackage[algo2e, noend, noline]{algorithm2e}
\usepackage{subcaption}

\providecommand{\DontPrintSemicolon}{\dontprintsemicolon}
\DontPrintSemicolon


\begin{document}

\CopyrightYear{2016} 
\setcopyright{acmcopyright}
\conferenceinfo{GECCO '16,}{July 20-24, 2016, Denver, CO, USA}
\isbn{978-1-4503-4206-3/16/07}\acmPrice{\$15.00}
\doi{http://dx.doi.org/10.1145/2908812.2908890}

\clubpenalty=10000 
\widowpenalty = 10000
\title{Convolution by Evolution}
\subtitle{Differentiable Pattern Producing Networks}
\numberofauthors{1}
\author{
\alignauthor
Chrisantha Fernando, Dylan Banarse, Malcolm Reynolds, Frederic Besse, David Pfau, Max Jaderberg, Marc Lanctot, Daan Wierstra\\
       \affaddr{Google DeepMind, London, UK}\\
       \email{chrisantha@google.com}
}

\maketitle
\begin{abstract}
In this work we introduce a differentiable version of the Compositional Pattern Producing Network, called the DPPN. Unlike a standard CPPN, the topology of a DPPN is evolved but the weights are learned. A Lamarckian algorithm, that combines evolution and learning, produces DPPNs to reconstruct an image. Our main result is that DPPNs can be evolved/trained to compress the weights of a denoising autoencoder from 157684 to roughly 200 parameters, while achieving a reconstruction accuracy comparable to a fully connected network with more than two orders of magnitude more parameters. The regularization ability of the DPPN allows it to rediscover (approximate) convolutional network architectures embedded within a fully connected architecture. Such convolutional architectures are the current state of the art for many computer vision applications, so it is satisfying that DPPNs are capable of discovering this structure rather than having to build it in by design. DPPNs exhibit better generalization when tested on the Omniglot dataset after being trained on MNIST, than directly encoded fully connected autoencoders. DPPNs are therefore a new framework for integrating learning and evolution.

\end{abstract}

%
%

%
%
\printccsdesc



\section{Introduction}

Compositional Pattern Producing Networks (CPPN) \cite{stanley2007compositional} were a major advance in evolutionary computation because they permitted evolution to efficiently optimize a model incrementally starting from a small number of parameters.

A CPPN is a very effective way of encoding a high dimensional output space with a low number of parameters, assuming some structure in the output space. This means that one can evolve CPPNs to represent images; as in Picbreeder~\cite{secretan2011picbreeder}, where a crowd of Internet users evolve images by selecting which CPPNs to breed.

The way CPPNs are currently optimized is by evolving the topology and the weights with NEAT~\cite{stanley2002evolving}. While effective for low dimensional output spaces, this process becomes inefficient for very large parameter spaces.

However, one can train neural networks with millions of parameters by exploiting gradient-based learning methods \cite{rumelhart1988learning}. The purpose of this work is to combine the success of gradient-based learning on neural networks with the capacity to optimize topologies provided by evolution. We call this new model the Differentiable Pattern Producing Network (DPPN). The DPPN works by Lamarckian evolution, i.e there is inheritance of acquired (learned) characteristics.

We show that using DPPNs we can rapidly reconstruct images by using fewer parameters than the number of pixels, and that DPPNs can be used in a HyperNEAT-like framework \cite{stanley2009hypercube} as an indirect encoding of a larger neural network. DPPNs also work in a Darwinian/Baldwinian framework \cite{santos2015phenotypic} in which the learned weights are not inherited directly, only the initial weights of the DPPN are inherited. However, the Lamarckian algorithm consistantly outperforms the other two variants.

The DPPN is more data efficient than the CPPN. In addition, it improves upon existing machine learning techniques by acting as a strong regularizer encouraging simpler solutions compared to those obtained by directly optimizing the weights of the larger network. For example we show that when a DPPN is trained using our algorithm to produce the 157684 parameters of a fully connected denoising autoencoder for MNIST digit reconstruction (MNIST is a standard benchmark for supervised learning consisting of labelled handwritten digits) \cite{vincent2008extracting}, it generates a convolutional architecture embedded within the fully connected feedforward network, in which each hidden unit contains a blob-like $28\times28$ weight matrix where the blob is smoothly moved over the receptive fields of hidden nodes. Evolution also discovers to crop and magnify the image. For example, a DPPN with only 187 parameters achieved a binary cross entropy (BCE) of 0.09 on the MNIST test set. Generalization to the Omniglot character set \cite{lake2015human} is also demonstrated to be superior to an equivalent directly encoded network.

\section{Background and Related Work}\label{sec:related}

The CPPN is a feedforward network which contains not only sigmoid and Gaussian functions but includes a wider set of transfer functions, for example periodic functions such as sine functions. CPPNs were invented by Ken Stanley \cite{stanley2007compositional} as an abstraction of natural development. CPPNs map a genotype coordinate set to a phenotype parameter set without local interaction between phenotypic elements, that is, each individual component of the phenotype is determined independently of every other component. The CPPN is in effect convolved over a set of coordinates to generate the output.

For example, a CPPN used to produce a square image of side length N would use an input coordinate set comprising $N\times N$ vectors of length 4; one vector datapoint for each pixel position in the $N\times N$ image to be produced. Typically, each datapoint is composed of the x,y coordinates, distance d(x,y) from the center of the image, and a fixed bias 1. The NxN datapoints are passed through the CPPN one by one, and the output image is generated by the CPPN sequentially, pixel by pixel. The CPPN's topology and weights are optimized by an evolutionary algorithm. A good example of this methodology is Picbreeder~\cite{secretan2011picbreeder}, in which a crowd of Internet users evolve images by selecting which CPPNs to breed.  Both the topology \textit{and} weights of the CPPN are evolved using mutation and crossover, starting with a minimal topology which grows nodes and edges. The NeuroEvolution of Augmented Topologies (NEAT) algorithm~\cite{stanley2002evolving} is used to constrain crossover to homologous parts of the CPPN, and to maintain topology diversity.

In HyperNEAT~\cite{stanley2009hypercube}, CPPNs are used as indirect compressed encodings of the weights of a larger neural network. The inputs to the CPPN are the coordinates of the presynaptic and postsynaptic neuron, and the output is the weight joining those two neurons. If a single CPPN must encode multiple layers of a deeper neural network then there are two possibilities, either an extra input is given signaling which layer of weights the CPPN is outputting~\cite{morse2013single}, or the CPPN is constrained to always have multiple output nodes, with a specific node outputting the weight for its assigned layer~\cite{pugh2013evolving}.

A limitation of the CPPN approach is that weights are evolved rather than being learned by gradient-based methods that utilize backpropagation. Such methods scale better than evolutionary methods with respect to the number of parameters in the model. They are able to optimize millions of parameters at once, e.g. for convolutional neural networks for performing object classification on ImageNet \cite{krizhevsky2012imagenet}. CPPNs and convolutional neural networks have previously been studied with CPPNs being used to evolve adversarial examples for convolutional network classifiers on ImageNet \cite{nguyen2014deep}. However, in that work the CPPN is not modified by gradient descent.

Convolutional neural networks \cite{fukushima1980neocognitron, lecun1989backpropagation} have made great strides in recent years in practical performance \cite{krizhevsky2012imagenet}, to the point where they are now a critical component of many of the best systems for challenging artificial intelligence tasks \cite{he2015deep, chen2014semantic, mnih2015human, kaiser2015neural}. These architectures were historically inspired by the structure of the mammalian visual system, namely the hierarchical structure of the ventral stream for visual processing \cite{felleman1991distributed} and the local, tiled nature of ``receptive fields" in the primary visual cortex \cite{hubel1962receptive}.

The engineering success of convolutional neural networks relative to fully connected neural networks is largely due to the strong regularization imposed by the convolutional structure: with far fewer weights to learn, the networks generalize better with less data. This architecture places strong prior assumptions on the data - namely that they are translation-invariant - and in most applications the architecture is decided on by the model designer rather than being automatically driven by the data.

It has also been shown that even greater improvements in the compression of neural network weights should be possible - even after removing most of the weights from the filters of a trained convolutional neural network, it is possible to predict the missing weights with high accuracy \cite{denil2013predicting}. This allows compression of the weights of convolutional neural networks in order to make them computationally more efficient \cite{Yang2015,jaderberg2014speeding}. It is of interest whether the appropriate simplifying structures can be {\em discovered} rather than designed, much like how evolution stumbled upon such structure for the mammalian visual system.

Recent work applied CPPNs in the HyperNEAT framework to evolve the weights of the 5 layer LeNet-5 convolutional neural network for MNIST character recognition \cite{verbancsics2013generative}. Classification performance with HyperNEAT alone used to optimize the weights of this network was very poor after 2500 generations, with only 50\% correct classifications. When HyperNEAT was used to initialize the weights of LeNet-5, prior to several epochs of gradient descent learning, correct classifications increased to 90\%. However, error rates of 0.8\% were obtained with backpropagation alone \cite{lecun1998gradient}. Also, there is no reduction in the number of parameters required to represent the resultant network because backpropagation is applied to the full convolutional network and not to the CPPN itself.

Previous work exists in evolving the topology of neural networks for learning. For example Bayer et al evolved the topology of the cells of an LSTM (Long Short Term Memory) recurrent neural network for sequence learning \cite{bayer2009evolving}, and more recently Jozefowicz et al explored the topology of LSTMs and GRUs improving on both \cite{zaremba2015empirical}.

\section{Methods}
Here we will begin by introducing the DPPN. We then describe the overall algorithm for optimizing a DPPN which consists of an evolutionary part which contains a learning part in its inner loop.

\subsection{DPPNs}
The DPPN is a modified implementation of a CPPN that can compute the gradients of functions with respect to the weights. A CPPN is a function $d$ that maps a coordinate vector $\vec{c}$ to a vector of output values $\vec{p} = (p_1,p_2,\ldots,p_n)$. The function is defined as a directed acyclic graph $\mathcal{G} = \{\mathcal{N},\mathcal{E}\}$ where $\mathcal{N}$ is a set of nodes and $\mathcal{E}$ is a set of edges between nodes. The set of input and output nodes are fixed - one for each dimension of the coordinate and output vectors respectively. Each node $n_i \in \mathcal{N}$ has a set of input edges $\mathcal{E}_i$ that can be changed by evolution, and a transfer function $\sigma_i \in \Sigma$ from a fixed list of nonlinearities associated with it. Each edge $e_j \in \mathcal{E}$ has a weight $w_j$ as well as input and output nodes $n^{in}_j, n^{out}_j$. The activation $a_i$ at node $n_i$ is given by $\sigma_i(\sum_{e_j \in \mathcal{E}_i} w_j a^{in}_j)$ -- the weighted sum over activations from input nodes passed through an activation function. The output values are simply the activations of the output nodes.

\begin{figure}[t]
\centering
\includegraphics[width=90mm]{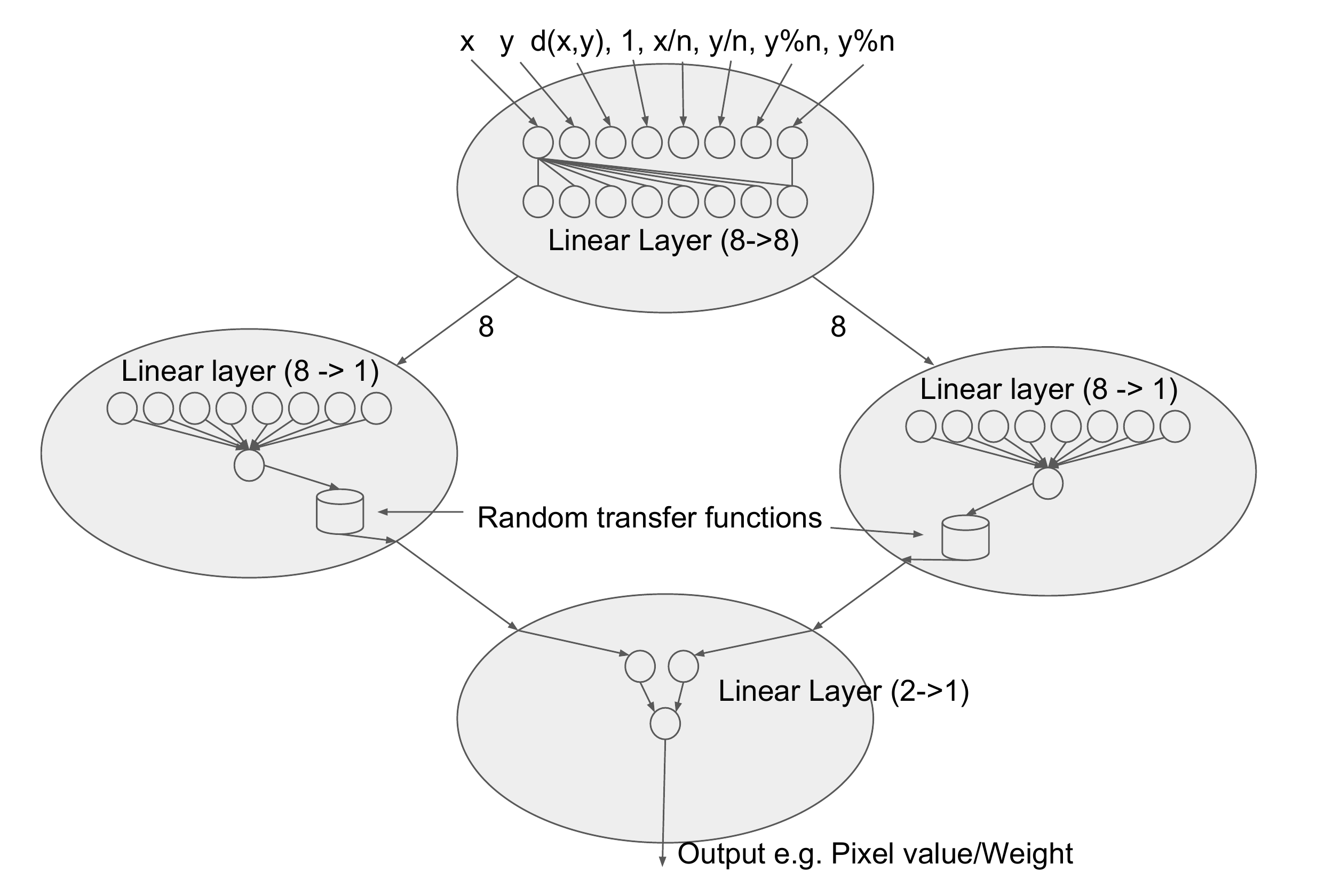}
\caption{Initial topology of the DPPN consists of two randomly chosen hidden units and an input and output node. Transfer functions are gaussian, sin, abs, ReLU, tanh, sigmoid, identity. An input node with a fully connected linear layer is shown, but we also use the more standard identity function for the input node with comparable results.}
\label{fig:1}
\end{figure}

For the DPPN, the node types used are as in previous CPPN papers \cite{stanley2007compositional}, i.e. sigmoid, tanh, absolute value, Gaussian ($e^{-x^2/2}$), identity and sine, plus rectified linear units (ReLU): $\sigma(x) = \mathrm{max}(x,0)$. We experiment with two kinds of input node, an identity node (as normally used in a CPPN) and a fully connected linear layer mapping $\vec{c}$ to a vector of equal dimensionality. There are no parameters in a node other than the weights and biases of its linear layer. The transfer functions all have fixed unlearnable parameters. Each DPPN is initialized with the topology shown in Figure~\ref{fig:1} with two random hidden units. We also experiment with complex initializations of fully connected feedforward DPPNs ranging from 5 to 15 initial nodes. The DPPN is encoded genetically as a connection matrix and a node list.

\subsection{Denoising Autoencoders}
A denoising autoencoder \cite{vincent2008extracting} is an unsupervised learning framework for neural networks. The network takes as input a noisy version $\widetilde{x}$ of the training data $x$, passes it through a set of layers $f_{\theta}(\widetilde{x}) = f^n_{\theta_n}(f^{n-1}_{\theta_n-1}(\ldots f^1_{\theta_1}(\widetilde{x})\ldots ))$ with parameters $\theta = \{\theta_1,\ldots,\theta_n\}$ and computes a loss $\ell(x,f_\theta(\widetilde{x}))$ between the noiseless data and the prediction of the network. In our experiments we use the mean squared error and the binary cross entropy. The loss function is usually minimized by some variant of gradient descent. In the DPPN, the output parameters $p$ are directly mapped into the parameters $\theta$ of the denoising autoencoder.

\subsection{Optimisation Algorithm}
A generic evolutionary algorithm can be used in the outer loop of the optimization algorithm. Learning takes place in each fitness evaluation just before evaluating the fitness function; a number of steps of gradient-based learning are performed starting from the inherited weights. In the Lamarckian version the learned weights are inherited (after mutation) by the offspring. In the Darwinian version the learned weights are discarded, and the initial weights of the parents are inherited (after mutation) by the offspring. We now describe the evolutionary algorithm, and the embedded learning algorithm in more detail.

\begin{algorithm2e}[t]
\small
\SetKwInOut{Input}{input}\SetKwInOut{Output}{output}
\Input{$P$ -- population of DPPNs}
\;
{\bf function} \textsc{GetFitness}$($DPPN $d)$ \;
\For{1000 steps}{
	Parameters $\vec{p} \leftarrow d(\vec{c})$ \;
	Copy $\vec{p}$ into denoising autoencoder parameters $\theta$ \;
	Choose minibatch $x$ of MNIST images \;
	Generate noisy minibatch $\widetilde{x} $\;
	Gradients $g_i \leftarrow \frac{\partial \ell(x,f_\theta(\widetilde{x}))}{\partial w_i}$ wrt DPPN weights \;
	Follow Adam update to DPPN weights $\{w_i\}$ using $\{g_i\}$\;
}
{\bf return } fitness = -MSE for 1000 MNIST training images
\;
{\bf function} \textsc{Main}() \;
\For{1000 tournaments}{
Choose two DPPN, $d_1, d_2 \in P$ \;
$(f_1, f_2) \leftarrow \textsc{GetFitness}(d_1, d_2)$ \;
Choose winner $A$ and loser $B$ \;
$B \leftarrow$ Mutate(Crossover($A$,$B$)) \;
}
\;
\caption{DPPN Trainer}\label{alg}
\end{algorithm2e}
\subsubsection{The Evolutionary Algorithm}
Two different evolutionary algorithms were used. The simpler evolutionary algorithm is the microbial genetic algorithm (mGA) \cite{harvey2011microbial} with a population size of 50. Two random agents are chosen, their weights are trained (see next section), each agent's fitness is evaluated, and a mutated copy of the winner overwrites the loser. There may be some probability of crossover, in which case the loser is parent B and the winner is parent A (see section on crossover). The second genetic algorithm is an asynchronous binary tournament selection algorithm running in parallel. This is identical to the mGA except that whenever more than two workers return a fitness, random pairs are chosen to undergo binary tournaments. This setup is used for the MNIST and Omniglot experiments which are computationally more demanding. The fitness of the DPPN is the negative loss.

\subsubsection{Mutation and Crossover Operators}
Three types of topology mutation are applied: add random node, remove random edge, add random edge. When a node is added, a random input node and a random output node are chosen and the new node is connected between them, care being taken to maintain the feedforward property of the DPPN. The initial weights to and from the new node are drawn from the same distribution as the weights of the initial DPPN.  The probability of node addition, edge addition and removal are typically, 0.3, 0.5, and 0.5 respectively per replication event. We also experiment with applying Cauchy mutation to the copy of the winner after fitness evaluation with a multiplicative co-efficient of 0.001 \cite{yao1999evolutionary}. Cauchy mutation is preferred because most mutations are small, but because of its heavy tail, a few are big, allowing escape from local optima.

The crossover operator is a merge in which the hidden units of both parents are combined, the input and output node of parent B is discarded, the input unit of parent A is connected to all the hidden units of parent B with random weights, and all the hidden units of parent B are connected to the output unit of parent A by random weights. Thus, each crossover results in an approximate doubling of the DPPN. No attempt is made as in NEAT to use innovation numbers. After crossover, a topological sort algorithm is used to reorder the connection matrix to make it upper-right triangular to enforce and check the feedforward property.

\subsubsection{The Learning Algorithm}
The learning phase embedded in the fitness evaluation cycle of an agent consists of 1000 steps of training carried out with a minibatch size of 32 data points into the DPPN. Gradients of the loss with respect to the CPPN weights are computed by backpropagation, which first computes the gradients of the loss with respect to the parameters, and then passes these backwards through the CPPN to be combined with the gradients of the parameters with respect to the CPPN weights: $\frac{\partial \ell(x,f_\theta(\widetilde{x}))}{\partial \vec{w}} = \frac{\partial \ell(x,f_\theta(\widetilde{x}))}{\partial \vec{p}}\frac{\partial \vec{p}}{\partial \vec{w}}$. For modifying weights of the DPPN we use Adam (adaptive moment estimation)  \cite{kingma2014adam} which is a momentum-based flavor of SGD that adaptively computes individual learning rates for each parameter by keeping an estimate of the first and second moments of the gradients. Two hyper-parameters, $\beta_1$ and $\beta_2$, are used to control the decay rates of two moving averages, $m_t$ for the gradient and $v_t$ for the squared gradient. These moving averages are then bias-corrected, resulting in an estimate of the moments $\hat{m_t}$ and $\hat{v_t}$. This algorithm is well suited for problems that incorporate a large number of parameters, as it is memory and computationally efficient. It combines the advantages of two popular methods: AdaGrad \cite{duchi2011adaptive}, which behaves well in presence of sparse gradients, and RMSProp \cite{graves2013generating}, which is able to cope with non-stationary objectives.

\subsection{Experiments}
In this section we describe experiments on image reconstruction, character denoising, compression ratios, and generalization from MNIST to Omniglot.

\subsubsection{Image Reconstruction Experiments}
A single randomly chosen $28\times28$ MNIST digit is chosen to be reconstructed. This is a simple benchmark task used to test various hyperparameter settings of the DPPN. The input batch to the DPPN is a $28\times28$ matrix of length 8 vectors. Each vector is constructed as
\[(x, y, \sqrt{x^2 + y^2}, 1, \frac{x}{N}, \frac{y}{N}, x~\mathbf{mod}~N, y~\mathbf{mod}~N), \mbox{ with } \]
$x$ and $y$ normalized to values in $[-1, 1]$, sampled in evenly spaced steps over the image, and with the target output for each input data point being the normalized pixel value at that $(x,y)$ location of the image. $N$ is a number encoded by the genotype of the DPPN. The fitness of a DPPN is the negative MSE on the entire $28\times 28$ image. One evolutionary run consists of a 1000 binary tournaments, after which the best fit agent is chosen and the best MSE reported.

\subsubsection{Denoising of MNIST images with a\\convolutional autoencoder}
The task is to reconstruct MNIST digits after 10\% of pixels are set to zero in the image. The convolutional network has an encoding layer with ReLU activation functions and a decoding layer with ReLU activation functions, with two kernels $7\times7$ in each layer, a stride of 2 and no pooling. The total number of parameters in this network is 202. For this experiment a DPPN with 6 output nodes is used to encode the weights of the kernels and the biases of the convNet. The first four outputs of the DPPN encode respectively: the weights of the first encoding kernel, the second encoding kernel, the first decoding kernel, and the second decoding kernel.The final two outputs encode biases. The input vector of each data point into the DPPN is 4 in length and encodes [x, y, $\sqrt{x^2 + y^2}$, 1].  $7\times7$  data points of length 4 are input into the DPPN, corresponding to $7\times7$  outputs of length 6. These are interpreted as the weights of the convolutional autoencoder. Up to 10000 fitness evaluations are carried out, each evaluation corresponding to the presentation of 32000 MNIST images from the training set. The final performance of a run is the MSE on a test set of 1000 MNIST images.

\subsubsection{Indirect encoding of a fully connected network}
\begin{figure}[t]
\centering
\includegraphics[width=85mm]{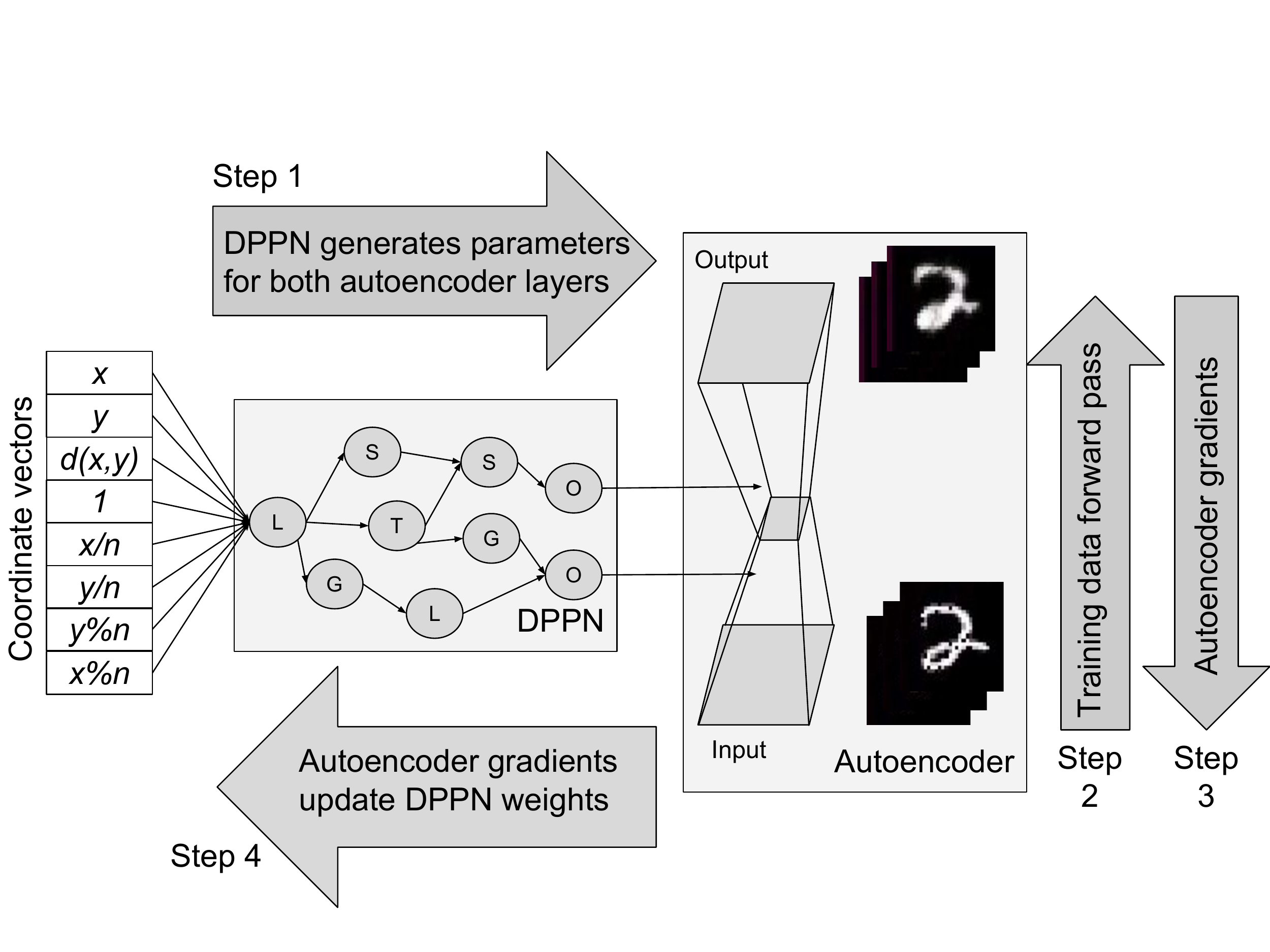}
\caption{The dual evolution and learning framework for training a DPPN based autoencoder.}
\label{fig:alg}
\end{figure}

The task is to reconstruct MNIST digits after 10\% of pixels are set to zero in the image. Figure~\ref{fig:alg} shows the logic of the training. We learn to indirectly encode a fully connected feedforward denoising autoencoder with one encoding layer with sigmoid activation functions and one decoding layer with sigmoid activation functions. The hidden layer has 100 units which are arranged on a $10\times10$ grid. Thus there are $28\times28\times10\times10\times2 + 28\times28 + 10\times10 = 157684$ parameters (weights and biases) which is three orders of magnitude more parameters than the convNet which performs the same task. The DPPN which encodes these parameters has two outputs, one for the encoding layer and one for the decoding layer. To obtain these parameters we passed 157684 input vectors into the DPPN each of length 8 which encode the following properties of the autoencoder:
\[(\text{x}\textsuperscript{in},\text{y}\textsuperscript{in}, \text{x}\textsuperscript{out},\text{y}\textsuperscript{out}, \text{D}\textsuperscript{in}, \text{D}\textsuperscript{out}, \text{layer},  1),\] where \text{x}\textsuperscript{in} and \text{y}\textsuperscript{in} are coordinates of the input neuron, \text{x}\textsuperscript{out} and \text{y}\textsuperscript{out} are the coordinates of the output neuron, and \text{D}\textsuperscript{in}, and  \text{D}\textsuperscript{out} are distances from the center of the input and output neuronal grids respectively. This produces $157684\times2$ parameters as output, but only the first $28\times28\times10\times10 + 10\times10$ elements of the first row and the second $28\times28\times10\times10 + 28\times28$ elements of the second row are used to encode the parameters of the autoencoder. A similar process consisting of a forwards pass through the DPPN, a copy of the DPPN outputs to the autoencoder, a forward and a backwards pass through the autoencoder with a MNIST minibatch, followed by backpropagation of these gradients through the DPPN is iterated 1000 times per fitness evaluation. After this training, the fitness of the DPPN is defined as the negative BCE (or MSE) over 1000 random MNIST images from the training set. The final loss of the run is the BCE (or MSE) on a test set of 1000 random MNIIST images.

\section{Results}

In this section, we present the experimental results.

\subsection{Image Reconstruction Experiments}
\begin{figure*}[t]
\centering
\includegraphics[width=155mm]{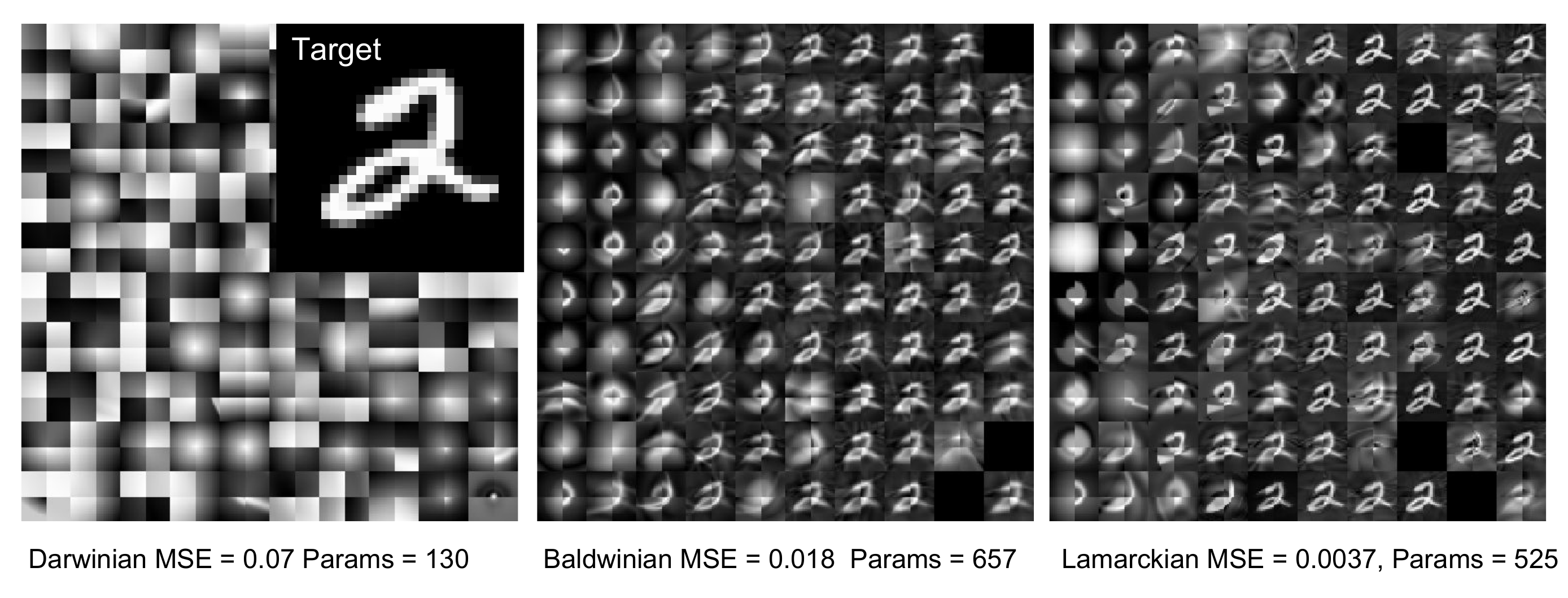}
\caption{Image reconstruction of the same handwritten digit 2 with Lamarckian, Baldwinian, and Darwinian inheritance of weights. The insert target image shows the character to reconstruct. The grids show the reconstructions produced during  evolutionary runs of 1000 tournaments, sampled every 10 tournaments, starting on the top left and proceeding to the bottom right corner. Lamarckian is better than Baldwinian is better than Darwinian. }
\label{fig:2}
\end{figure*}

Figure~\ref{fig:2}(right) shows the details of an evolutionary run in which a population of 50 DPPNs, with crossover probability of 0.2, initialized with 4 node DPPNs, is evolved to reconstruct the handwritten digit 2, evolved with the full Lamarckian algorithm, i.e. where learned weights are inherited by the offspring. Figure~\ref{fig:2}(middle)  shows the same setup but with Baldwinian evolution, i.e. where learning takes place but where there is no inheritance of acquired characteristics. Finally Figure~\ref{fig:2}(left) shows the same setup with pure Darwinian evolution with Cauchy mutation of weights with a co-efficient of 0.001 in which there is no learning of weights at all. This final setting is the closest to a CPPN. In the examples shown Lamarckian inheritance achieves a MSE of 0.0036, Baldwinian 0.02 and Darwinian 0.12.

Batch runs of size 10 (without crossover) show Lamarckian learning to have a mean MSE of $0.021~(\pm 0.006)$\footnote{All intervals in this paper represent 95\% confidence intervals.}, compared to Baldwinian runs which show a mean MSE of $0.037~(\pm 0.006)$ and Darwinian runs which show a mean MSE of $0.079~(\pm 0.006)$. We therefore conclude that for this task, Lamarckian is more effective than Baldwinian which is more effective than Darwinian. Traditionally CPPNs were initialized with minimal networks. We find it is also effective to initialize DPPNs with larger networks (5 to 10 hidden units) which are fully connected in the upper right triangle of the connection matrix. Batch runs show a mean MSE of $0.02~(\pm 0.006)$ starting large, compared to a mean MSE of $0.021~(\pm 0.006$) starting small, showing no significant difference. We also tried a hybrid variant with learning rates and Cauchy mutation. There was a small non-significant benefit to adding Cauchy noise for all learning rates investigated, therefore in the later runs we used a Cauchy mutation coefficient of 0.0001. Additive bloat punishments produced no improvement at any level, and produced a significantly worse MSE when greater than $0.001(n + e)$, where $n$ and $e$ are the number of nodes and edges in the DPPN. 1000 steps of learning produced a mean MSE of $0.026~(\pm 0.006)$ compared to a 100 steps of learning which produced a mean MSE of $0.035~(\pm 0.006)$, so more learning is better.

\subsection{The Effect of Crossover}
\begin{figure}[t]
\centering
\begin{subfigure}{.2\textwidth}
\includegraphics[width=35mm]{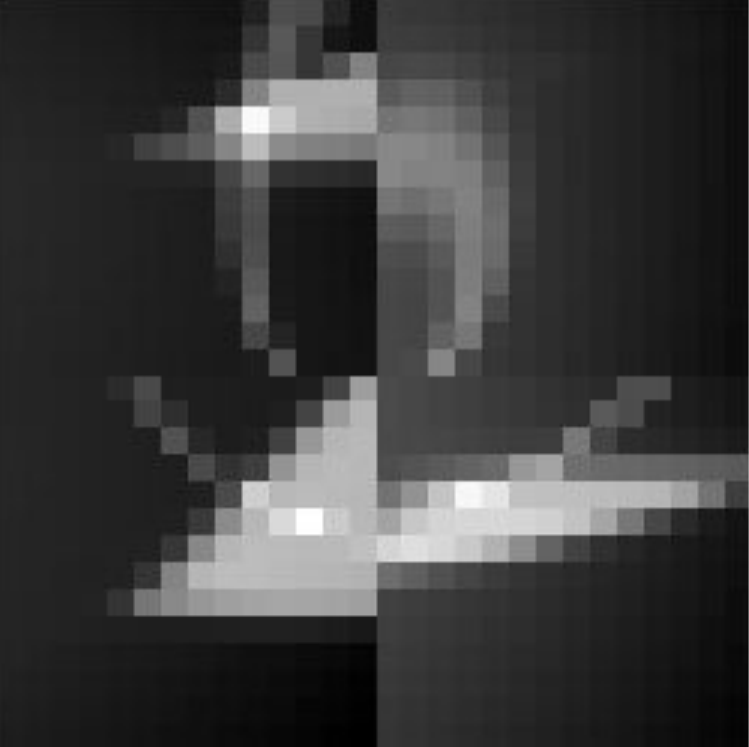}
\end{subfigure}
\begin{subfigure}{.2\textwidth}
\includegraphics[width=35mm]{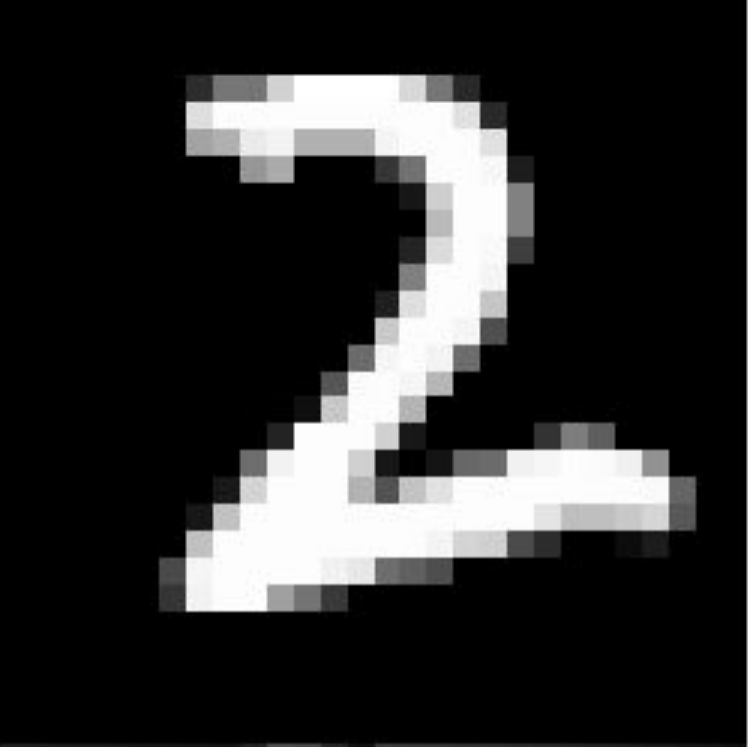}
\end{subfigure}
\caption{Without crossover the image reconstruction is more messy and has a higher MSE of 0.03 after 1000 tournaments (left), compared to an MSE of 0.003 with crossover (right). Number of parameters without crossover = 112}
\label{fig:4}
\end{figure}

Figure~\ref{fig:4} shows the same setup as a Figure~\ref{fig:2}(right) run but without crossover. There is an order of magnitude difference in MSE, 0.003 with crossover compared to 0.03 without crossover. The reconstruction is qualitatively worse without crossover. Batch runs of size 10 show an order of magnitude benefit of crossover, with MSE of $0.005~(\pm 0.001)$ with crossover probability 0.2, compared to a MSE of $0.021~(\pm 0.006)$, without crossover. A trivial reason for the benefit of crossover may be that it merely increases the size of the networks (717 compared to 112 parameters) so allowing a greater number of parameters to be optimized by gradient descent, possibly reducing the chance of getting stuck in a local optimum. Another factor is that merging DPPNs allows informational merging of different useful parts of the image reconstructed by different individuals in the population.

\subsection{Can the DPPN efficiently compress the\\weights of denoising autoencoders?}
\begin{figure}[t]
\centering
\includegraphics[width=70mm]{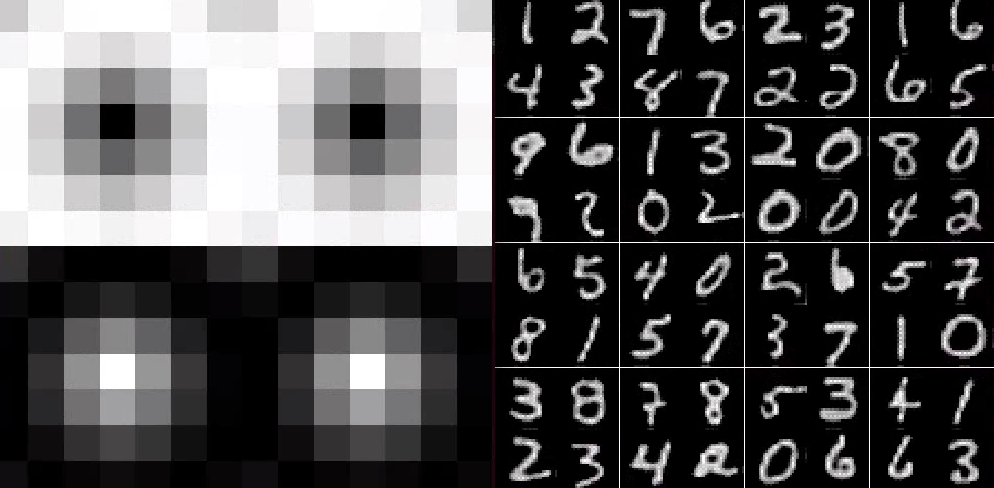}
\caption{DPPN produced encoding and decoding kernels for a convolutional denoising autoencoder on MNIST. Left: Encoding and decoding kernels. Right: Digit reconstructions}
\label{fig:6}
\end{figure}

Figure~\ref{fig:6}  shows the 2 encoding (left column) and 2 decoding kernels (right column) evolved by the DPPN for the convolutional denoising autoencoder, along with the digit reconstructions and fitness graph showing that 1000 tournaments are sufficient for an MSE of 0.01 on the test set. The DPPN discovers regular on-center and off-center receptive fields resembling those of retinal ganglion cells for image smoothing which removes most of the uncorrelated dropout noise from the reconstruction.

\begin{figure}[t]
\centering
\includegraphics[width=70mm]{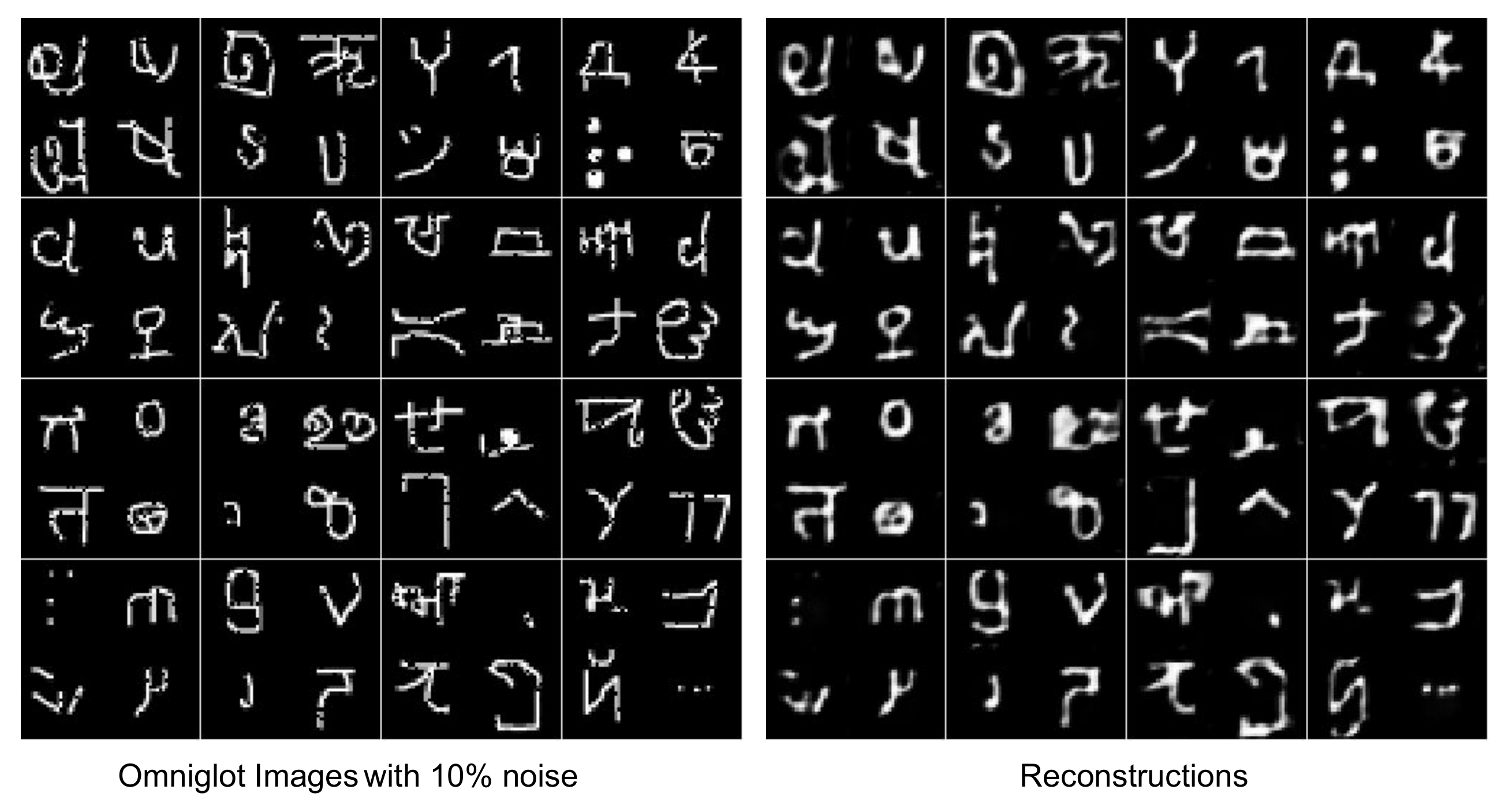}
\caption{Omniglot reconstructions by a 191 parameter network trained only on MNIST(BCE = 0.096) achieves a BCE of 0.121.}
\label{fig:omniglot}
\end{figure}

\begin{figure}[t!]
\centering
\includegraphics[width=80mm]{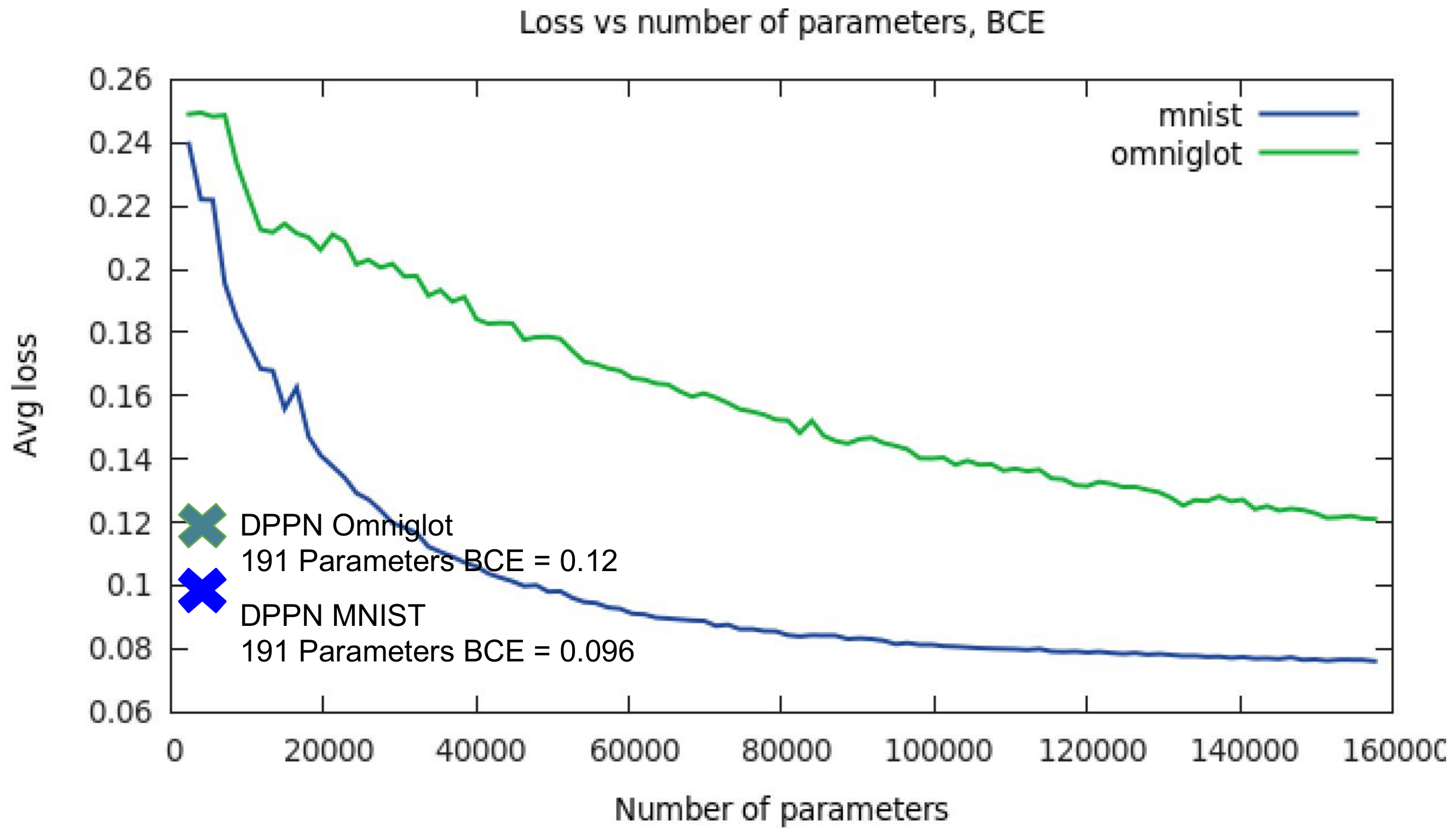}
\caption{The loss of directly encoded networks with hidden layer sizes ranging from 1 to 100 nodes are compared with the DPPN encoded 100 Hidden node network. }
\label{fig:9}
\end{figure}

\begin{figure}[t!]
\centering
\includegraphics[width=90mm]{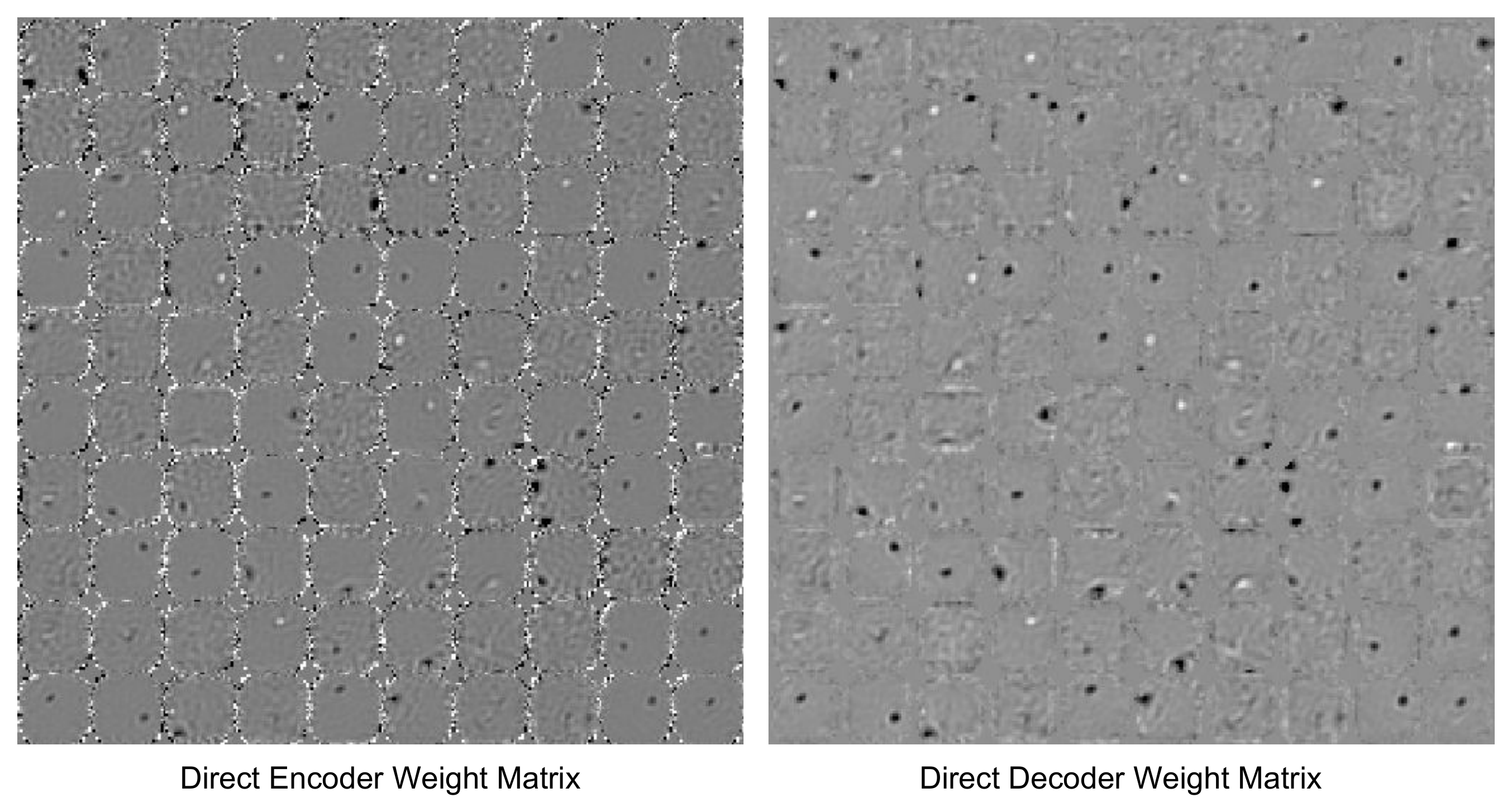}
\caption{The encoding and decoding weight matrices of a fully connected 100 hidden unit denoising autoencoder trained directly are much less regular. This network was trained with MSE criterion for 500 epochs ($500\times50000$ samples). The network contained $157684$ parameters. It achieved an BCE of 0.0761 on MNIST and generalized with BCE of 0.121 to Omniglot.}
\label{fig:8}
\end{figure}

\begin{figure*}[t!]
\centering
\includegraphics[width=100mm]{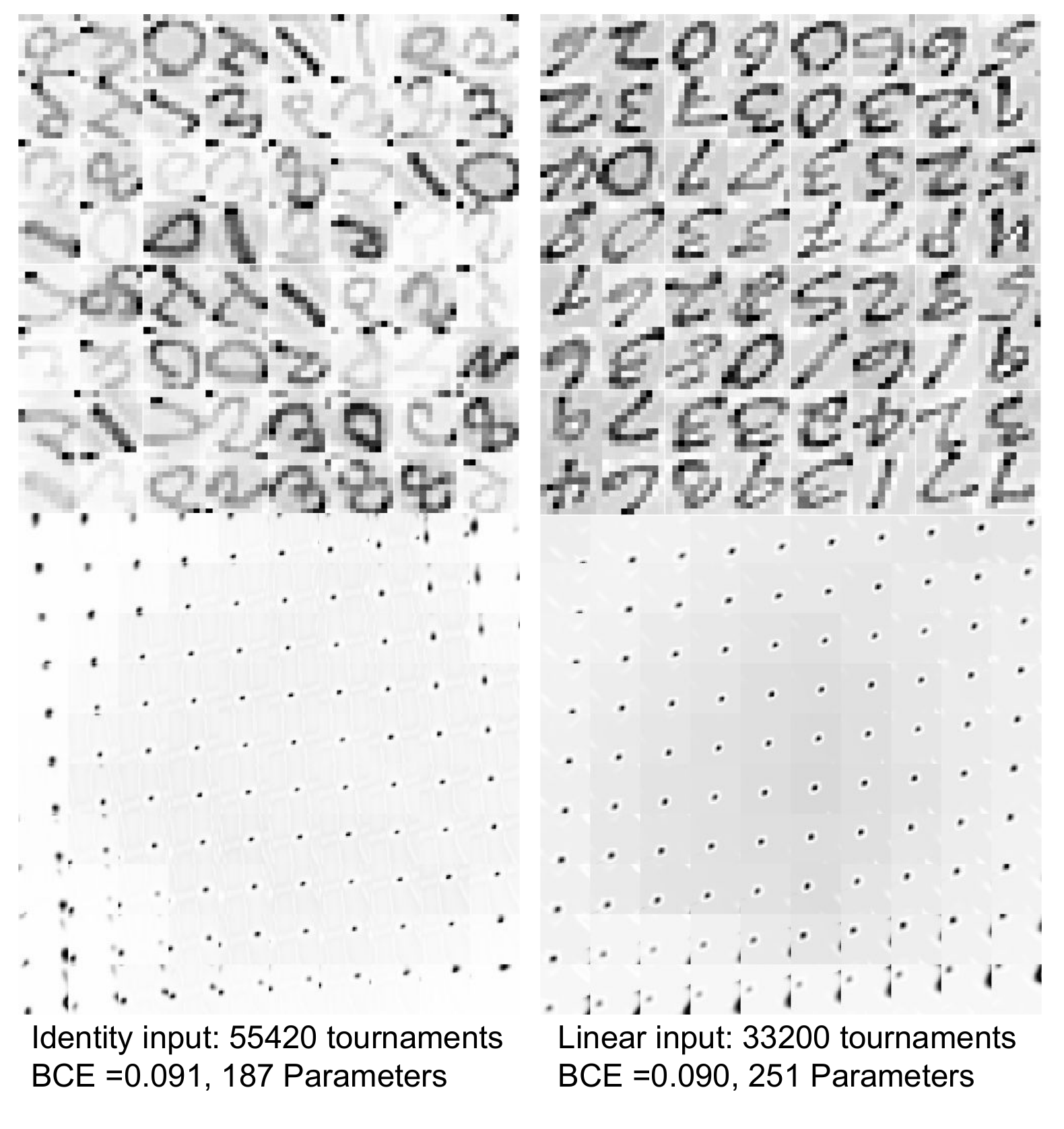}
\caption{$10\times10$ representations of MNIST digits in the hidden layer of the denoising autoencoder (top) and corresponding encoding layer weights for the denoising autoencoder (bottom). Left shows a 187 parameter network with no linear layer at input, and Right shows a 251 parameter network with a fully connected linear layer at input. The hidden layer representations have been rotated, cropped and inverted by the encoder. }
\label{fig:10}
\end{figure*}

\begin{figure*}[t!]
\centering
\includegraphics[width=160mm]{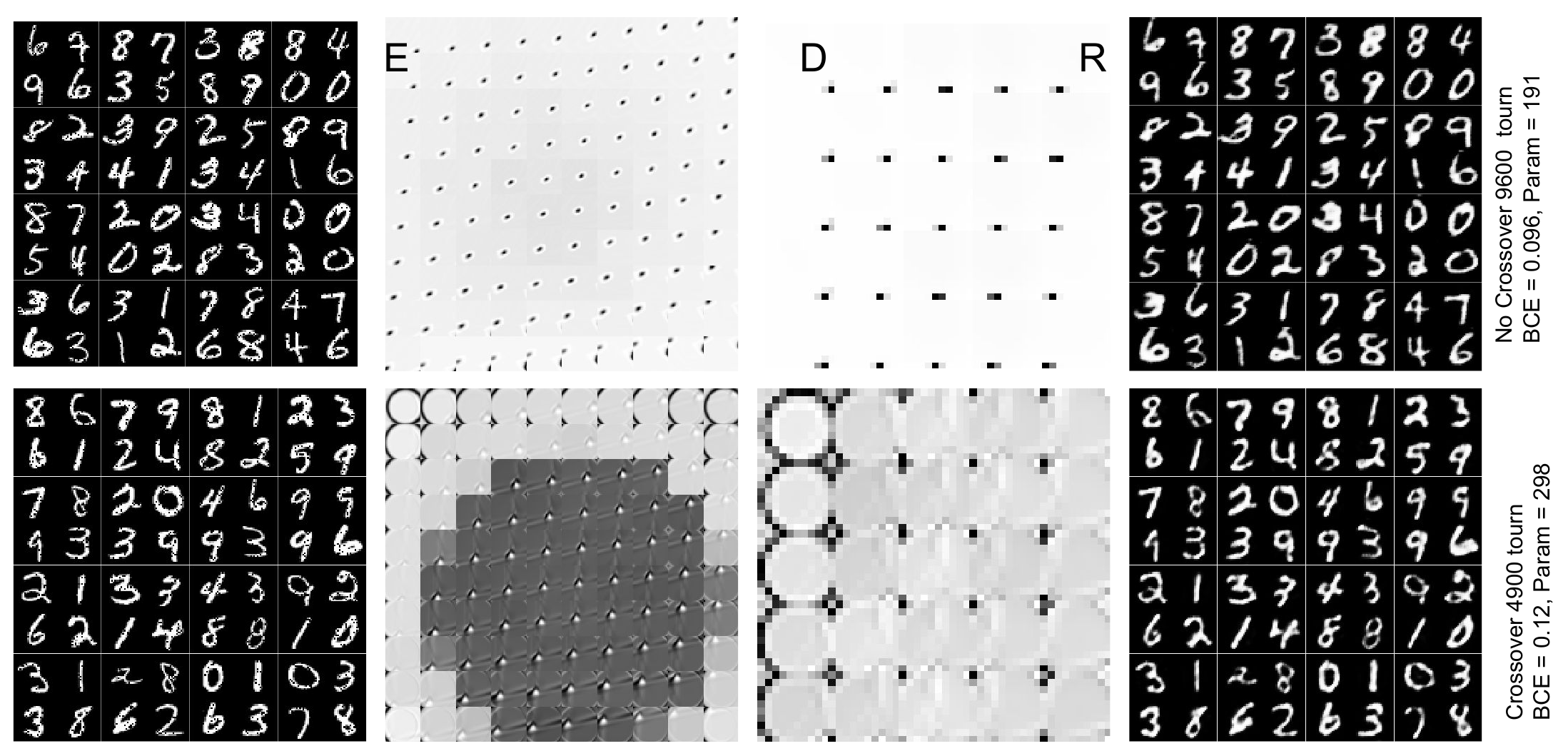}
\caption{Discovery of convolutional filters in a fully connected denoising autoencoder. The input digits are shown on the left, followed by the 10 x 10 neuron encoding layers weight matrix (E) and a 5 x 5 sample of the 28 x 28 neuron decoding layers weight matrix (D), and finally the reconstructions (R) on the right. Note the highly regular weight matrices of the E and D layers compared to the directly encoded E and D matrices in the next figure.}
\label{fig:7}
\end{figure*}

Figure~\ref{fig:7} compares the performance of a DPPN with (top) and without crossover (bottom) on producing the $157684$ parameters of a fully connected denoising autoencoder. In both cases the DPPN rediscovers convolutions by learning the on/off center kernels and then convolving them over the $28\times28$ receptive fields of the 100 hidden units. The decoding $28\times28$ layer also discovers this convolutional structure in a fully connected network. This is in contrast to the receptive fields normally learned by such networks which are much less regular, see Figure~\ref{fig:8}. The extent of effective compression achieved of the autoencoder's parameters by the DPPN is remarkable, see Figure~\ref{fig:9}, which shows that the DPPN encoded network can achieve much lower BCE (0.096) than a directly encoded network with the same number of parameters (> 0.24). Furthermore, it is capable of generalization to the Omniglot dataset with BCE of 0.121 which is better than an equivalent directly encoded network, see Figure~\ref{fig:omniglot}. A video in Supplementary material shows the evolution of MNIST reconstructions throughout a run.

\section{Discussion and Conclusion}
The results demonstrate that DPPNs and the associated learning algorithms described here are capable of massively compressing the parameters of larger neural networks, and improve upon the performance of CPPNs trained in a Darwinian manner. Because the hidden layer has a $10\times10$ grid structure, we can visualize the activations in the hidden layer for each digit, see Figure~\ref{fig:10} which shows the hidden layer activations of a fully connected denoising autoencoder encoded by a DPPN with an identity node as input vs. a DPPN with a fully connected linear node as input.   Both produce comparable BCEs with roughly the same number of parameters.

One of the advantages of this symbiosis between evolutionary and gradient-based learning is that it allows optimization to better avoid being stuck in local optima or saddle points. In the future, this framework holds potential for training much deeper neural networks and being applied to other learning paradigms.

\section{Acknowledgments}
Thanks to John Agapiou for help with the code, and Ken Stanley, Jason Yosinski, and Jeff Clune for useful discussions.

%
\bibliographystyle{abbrv}
\bibliography{sigproc}  
%
%

\end{document}